\documentclass[conference]{IEEEtran}
\IEEEoverridecommandlockouts
\usepackage{cite}
\usepackage{amsmath,amssymb,amsfonts}
\usepackage{url}
\usepackage{graphicx}
\usepackage{textcomp}
\usepackage{xcolor}
\usepackage{algorithm}
\usepackage{algpseudocode}
\usepackage{booktabs}
\usepackage{tabularx}

\def\BibTeX{{\rm B\kern-.05em{\sc i\kern-.025em b}\kern-.08em
    T\kern-.1667em\lower.7ex\hbox{E}\kern-.125emX}}
\begin{document}

\title{Through \textit{BrokenEyes}: How Eye Disorders Impact Face Detection?\\}

\author{\IEEEauthorblockN{Prottay Kumar Adhikary}
\IEEEauthorblockA{\textit{Indian Institute of Technology Delhi } \\
\textit{Delhi, India}\\
\url{}}
}

\maketitle

\begin{abstract}
Vision disorders significantly impact millions of lives, altering how visual information is processed and perceived. In this work, a computational framework was developed using the BrokenEyes system to simulate five common eye disorders: Age-related macular degeneration, cataract, glaucoma, refractive errors, and diabetic retinopathy and analyze their effects on neural-like feature representations in deep learning models. Leveraging a combination of human and non-human datasets, models trained under normal and disorder-specific conditions revealed critical disruptions in feature maps, particularly for cataract and glaucoma, which align with known neural processing challenges in these conditions. Evaluation metrics such as activation energy and cosine similarity quantified the severity of these distortions, providing insights into the interplay between degraded visual inputs and learned representations.

\end{abstract}


\section{Introduction}
Face detection is a fundamental visual task, yet its reliability depends on the integrity of the visual system. The human visual pathway reflects a tight coupling between ocular input and downstream neural processing.\cite{b1} Eye disorders such as refractive errors, cataract, glaucoma, age-related macular degeneration (AMD), and diabetic retinopathy are among the leading causes of visual impairment worldwide.\cite{b2} Each condition selectively alters the visual signal available to the brain, producing deficits that are relevant to face detection; for example, AMD predominantly impairs central vision and high-acuity tasks, whereas glaucoma constrains peripheral vision and reduces situational awareness.\cite{b3} These impairments extend beyond low-level sensory loss: reduced visual input has been shown to influence higher-order regions, including the fusiform face area (FFA), which is central to face processing.\cite{b4} Prolonged deprivation can also induce cortical reorganization or diminished activation, underscoring the interdependence between sensory input and neural representation.\cite{b5}

Prior work has examined disorder-specific effects on face perception. Studies of prosopometamorphopsia (PMO), for instance, indicate that facial distortions can occur without abolishing recognition, suggesting separable mechanisms for perception and identity processing.\cite{b6} Complementary research on visual adaptation further emphasizes the role of cortical dynamics in face perception.\cite{b7} These findings motivate computational models that explicitly simulate impaired inputs and evaluate their impact on learned representations.

\begin{figure}[t]
\centering
\includegraphics[width=\columnwidth]{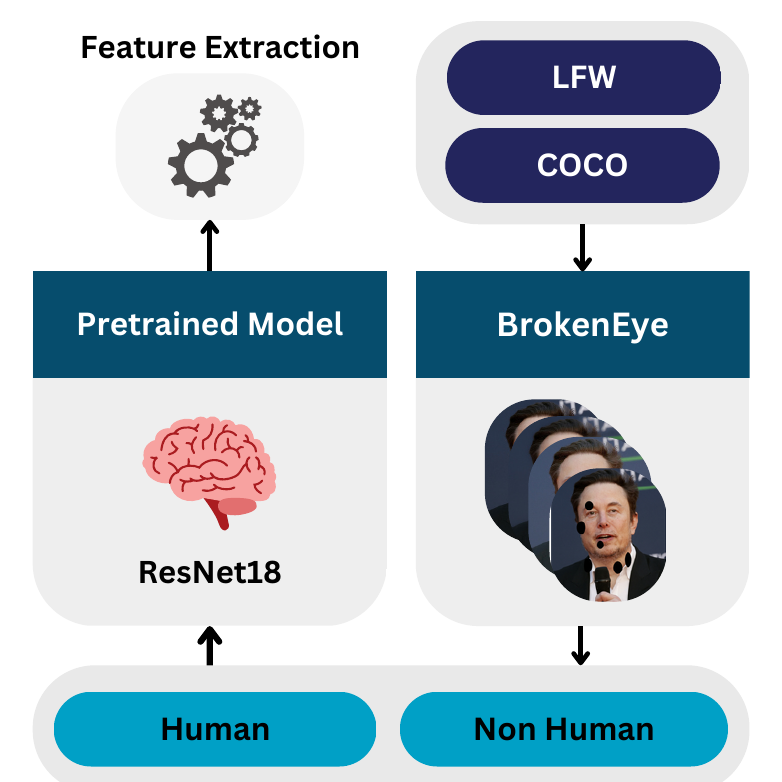}
\label{fig:pipeline}
\caption{End-to-end experimental pipeline. Raw images from LFW and MS-COCO are processed through the BrokenEyes filter generator to simulate five visual impairments, followed by disorder-specific fine-tuning of ResNet18. Feature maps extracted from layer4 are compared against the normal model using activation energy and cosine similarity to quantify representation drift.}
\end{figure}

\begin{figure*}[t]
  \includegraphics[width=\textwidth]{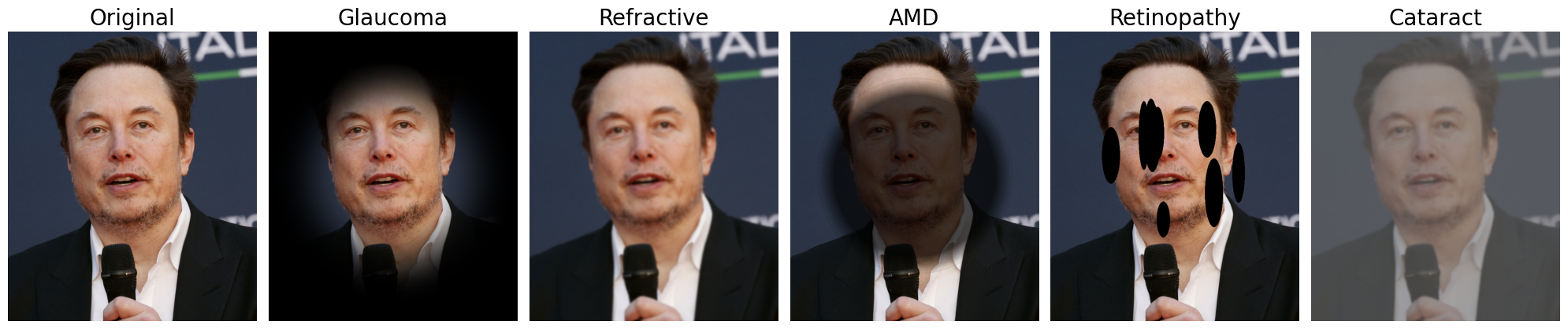}
  \label{fig:examples}
  \caption{Sample data from the dataset. The figure illustrates disorder-specific degradations (AMD, cataract, glaucoma, refractive error, and retinopathy) applied to identical inputs, highlighting characteristic changes such as central scotoma, haze/blur, peripheral vignetting, defocus, and scattered occlusions.}
\end{figure*}

To examine these effects computationally, this work simulates five common eye disorders and evaluates their impact on hierarchical feature representations in ResNet18,\cite{b8} a widely used convolutional neural network. Feature maps from models trained under normal and disorder-specific conditions are compared to assess how degraded inputs reshape internal representations. Using the proposed \textbf{BrokenEyes} filter generator, realistic visual filters are curated to approximate disorder-specific degradations, and model performance and feature alignment are quantified across conditions, as summarized in Figure 1. The main contributions of this work are: (1) a disorder-specific image degradation framework (\textbf{BrokenEyes}) that simulates five clinically common visual impairments, (2) a controlled experimental pipeline using human and non-human datasets to train disorder-aware ResNet18 models, and (3) a feature-level analysis that quantifies representation shifts via activation energy and cosine similarity.

\section{Dataset}
\label{data}
The dataset used in this study combines two widely known sources: the \textit{Labelled Faces in the Wild (LFW)}\cite{b9} dataset and the \textit{MS-COCO 2017}\footnote{https://cocodataset.org/} dataset. LFW provides unconstrained face images and was used to form the \textit{human} class. MS-COCO is a large-scale, general-purpose collection with diverse objects and scenes, and was used to form the \textit{non-human} class. To minimize bias and ensure consistent training conditions, a matched subset was curated from both datasets by filtering for sufficient resolution, removing obvious duplicates, and balancing the per-class sample counts. All images were then resized and normalized to match the input requirements of ResNet18. The final dataset supports binary classification (human vs. non-human) under normal vision and five disorder-specific degradations, enabling controlled evaluation of how visual impairments affect learned representations.

\subsection{BrokenEye: Filter Generation Framework}
To simulate the visual impairments associated with five common eye disorders, a novel filter generation system, named \textbf{BrokenEye}, has been proposed. This system applies realistic, disorder-specific degradations to images so that the resulting samples approximate the perceptual distortions reported clinically. The generated images represent different impairment mechanisms as shown in Figure 2 and are designed to preserve overall scene content while selectively altering the visual signal in ways that reflect each condition:

1. \textbf{Glaucoma Simulation:} Glaucoma is characterized by tunnel vision, which is simulated using a vignette effect. A black overlay with a circular transparent region at the center is applied, gradually fading into darkness toward the edges. The vignette is smoothed using Gaussian blurring, reproducing the peripheral vision loss typically associated with glaucoma.

2. \textbf{Refractive Errors Simulation:} Refractive errors are simulated using randomized Gaussian blur to mimic out-of-focus vision. The blurring intensity and kernel size are chosen randomly within a range to replicate varying degrees of nearsightedness, farsightedness, or astigmatism. This randomness adds natural variation, resembling the diverse presentations of refractive errors.

3. \textbf{AMD Simulation:} To replicate the central vision loss caused by age-related macular degeneration, a central darkened region with blurred edges is introduced. The overlay begins with an opaque central area, gradually decreasing in opacity toward the edges, mimicking the progressive loss of detail in the central visual field. The effect is further enhanced with a gradient blur to produce realistic \textit{central scotomas}.

4. \textbf{Retinopathy Simulation:} Diabetic retinopathy causes localized vision distortions, such as dark patches or floaters. This is simulated by adding random black elliptical shapes across the image. The positions, sizes, and counts of these patches are randomized to represent the scattered nature of retinal damage.

5. \textbf{Cataract Simulation}: The cataract filter applies Gaussian blurring and desaturation to simulate the haziness and reduced contrast characteristic of this condition. First, the image is desaturated by reducing its color saturation in the HSV color space, followed by the addition of a light haze. A Gaussian blur is then applied with a large kernel size, replicating the blurred vision experienced by individuals with cataracts.

Together, these filters produce images that closely resemble the visual experiences of individuals with these disorders, enabling the training and evaluation of models in conditions that approximate real-world impairment profiles while keeping the underlying semantic content stable.

\begin{table}[!h]
\caption{Human and non-human images across disorder conditions}
\centering
\renewcommand{\arraystretch}{1.15}
\begin{tabularx}{\columnwidth}{@{}l>{\raggedleft\arraybackslash}X>{\raggedleft\arraybackslash}X@{}}
\toprule
\textbf{Condition} & \textbf{Human} & \textbf{Non-Human} \\
\midrule
Normal             & 1414           & 1309              \\
AMD                & 1414           & 1309              \\
Cataract           & 1414           & 1309              \\
Glaucoma           & 1414           & 1309              \\
Retinopathy        & 1414           & 1309              \\
Refractive Error   & 1414           & 1309              \\
\midrule
\textbf{Total}     & \textbf{8484}  & \textbf{7854}     \\
\bottomrule
\end{tabularx}
\label{table:dataset_stats}
\end{table}

\subsection{Statistics}
The dataset generated after applying the filters to the original images ensures an equal distribution of samples across the six visual conditions (normal vision and five disorders). The counts per condition are summarized in Table~\ref{table:dataset_stats}. Overall, the dataset includes 8,484 human images and 7,854 non-human images, providing near-balanced classes and consistent per-condition coverage. This design enables controlled comparisons across disorders while maintaining stable class priors, which is important for interpreting differences in model behavior as a function of visual degradation rather than dataset imbalance. The resulting corpus supports robust training and evaluation of models under degraded vision, facilitating analysis relevant to computational neuroscience, vision science, and accessibility-oriented machine learning.

\section{Methodology}
\label{method}
We formulate a controlled transfer-learning pipeline to quantify how disorder-specific degradations reshape deep feature representations. ResNet18 is used as the backbone, and disorder-aware fine-tuning is performed on matched splits of human vs.\ non-human images across six visual conditions. Feature representations are extracted from the final convolutional stage and compared using cosine similarity and activation energy to isolate representation drift induced by each impairment.

Each dataset was stratified into training (70\%), validation (15\%), and testing (15\%) partitions with class balancing across normal vision, AMD, cataract, glaucoma, refractive errors, and diabetic retinopathy.

\vspace{0.3cm}
\begin{algorithm}
\caption{Disorder-Aware Training and Feature Comparison}
\label{alg:training}
\begin{algorithmic}[1]
\Require Filtered datasets $\{D_{n}, D_{a}, D_{c}, D_{g}, D_{ref}, D_{ret}\}$; pretrained ResNet18 backbone
\Ensure Disorder-specific models $\{M_d\}$ and feature metrics $\{\text{sim}_d, \text{energy}_d\}$
\State Split each $D_d$ into train/val/test (70/15/15)
\State Initialize ResNet18 with ImageNet weights; replace $\texttt{fc}$ with a binary classifier
\For{each disorder $d \in \{n, a, c, g, ref, ret\}$}
    \State Freeze backbone; train $\texttt{fc}$ for 5 epochs with Adam and NLL loss
    \State Unfreeze layer4; fine-tune for 1 epoch with reduced learning rate
    \State Save trained model $M_d$
\EndFor
\State Extract baseline feature map $F_n$ from layer4 of $M_n$
\For{each disorder $d \in \{a, c, g, ref, ret\}$}
    \State Extract feature map $F_d$ from layer4 of $M_d$
    \State $\text{sim}_d \leftarrow \cos(F_n, F_d)$
    \State $\text{energy}_d \leftarrow \sum |F_d|$
    \State Save metrics; visualize $F_n$ vs.\ $F_d$ heatmaps
\EndFor
\end{algorithmic}
\end{algorithm}

\vspace{0.3cm}
\subsection{Training the Human vs. Non-Human Classifier}
We fine-tune \textbf{ResNet18}\cite{b8} for binary classification by replacing the fully connected layer with a single linear head followed by log-softmax. Inputs are resized and normalized, and standard augmentation is applied. Training uses Adam (learning rate $10^{-3}$), negative log-likelihood loss, and early monitoring on the validation set to mitigate overfitting.

\subsection{Training Disorder-Specific Models}
For each disorder condition, a separate model is initialized from the same pretrained backbone and fine-tuned on the corresponding filtered dataset. The same optimization protocol is used to ensure comparability, and evaluation is performed on the held-out test split for each disorder.

\subsection{Feature Extraction and Comparison}
Feature maps are collected from layer4 (final convolutional block) to capture high-level, spatially localized representations. Given a fixed test image, we compute cosine similarity between the flattened feature maps of the normal model and each disorder-specific model. Activation energy is computed as the sum of absolute activations to measure response magnitude. Together, these metrics quantify representational alignment and intensity shifts induced by each impairment.\\

All models are implemented in PyTorch and trained on an NVIDIA A100 GPU with torchvision-based data loading and augmentation. Similarity and energy metrics are computed using tensor operations, and feature map differences are visualized with Matplotlib heatmaps. Code and dataset are available at: \url{github.com/proadhikary/BrokenEyes}

\begin{figure*}[!ht]
\label{fig:feature_map_heatmaps}
  \includegraphics[width=\textwidth]{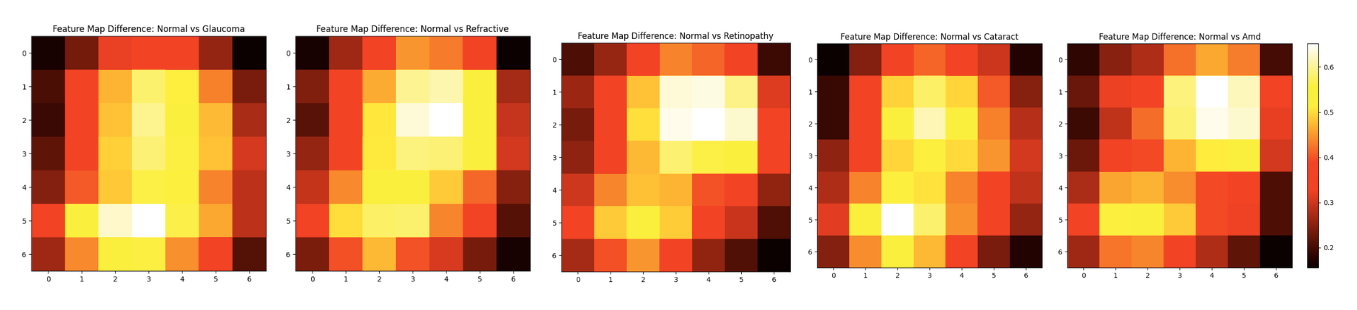}
  \caption{Feature map difference visualization. Each heatmap depicts the absolute deviation between the normal model's layer4 activation map and the corresponding disorder-specific model for the same input, highlighting spatial regions where representation drift is concentrated. Warmer colors indicate larger deviations, revealing that cataract and glaucoma produce the most widespread disruptions compared with AMD, refractive error, and retinopathy.}
\end{figure*}

\section{Result and Discussion}
\label{result}
The results from this computational analysis provide intriguing insights that could be linked to human neural processing and visual impairments. By analyzing how a deep learning model processes distorted images, we can draw parallels with how the human visual system might respond to similar conditions. Specifically, this study sheds light on the potential disruptions caused by different eye disorders on hierarchical feature representations in the brain, which are critical for tasks like face detection and object recognition.

The model trained on the normal dataset was first evaluated on the human vs.\ non-human classification task using the held-out test split (409 samples). The classifier achieved 100\% accuracy, correctly separating all samples into their respective classes under normal vision conditions. This establishes a clean baseline for downstream comparison, as it indicates that the backbone (ResNet18) with the specified fine-tuning protocol can reliably discriminate the two classes when visual inputs are undistorted. Figure 3 illustrates representative test samples, highlighting the diversity of poses, scales, and backgrounds. These baseline observations allow subsequent disorder-specific deviations to be interpreted as effects of input degradation rather than insufficient model capacity.

The disorder-specific models were evaluated on the same test split, and predictions were collected for images representing normal vision and the five simulated impairments. Table~\ref{table:evaluation_results} reports the predicted class confidence for the ``human'' category. While all models still predicted the correct class label, confidence scores systematically decreased under degraded inputs, indicating reduced separability of the learned feature embeddings. Notably, AMD and glaucoma show the largest confidence drops, consistent with their disruption of central or peripheral visual fields, which are important for global face structure and spatial context.

\begin{table}[h]
\caption{Evaluation: Confidence Scores (Human Class)}
\centering
\begin{tabularx}{\columnwidth}{@{}l>{\raggedleft\arraybackslash}X@{}}
\toprule
\textbf{Condition} & \textbf{Confidence Score} \\
\midrule
Normal             & 0.9394 \\
AMD                & 0.6795 \\
Cataract           & 0.7702 \\
Glaucoma           & 0.7060 \\
Refractive Error   & 0.8503 \\
Retinopathy        & 0.8609 \\
\bottomrule
\end{tabularx}
\label{table:evaluation_results}
\end{table}

The results indicate that AMD and glaucoma lead to lower confidence scores compared to normal vision and other disorders. This suggests that the classifier requires stronger evidence to sustain a positive ``human'' decision under these impairments, consistent with reduced signal quality in features that capture facial configuration.

\begin{table}[h]
\caption{Feature Map Comparison}
\centering
\begin{tabularx}{\columnwidth}{@{}l>{\raggedleft\arraybackslash}X>{\raggedleft\arraybackslash}X@{}}
\toprule
\textbf{Disorder}  & \textbf{Activation Energy} & \textbf{Cosine Similarity} \\
\midrule
AMD               & 24294.71                    & 0.9344 \\
Cataract          & 30180.73                    & 0.6350 \\
Glaucoma          & 29372.11                    & 0.4551 \\
Refractive Error  & 25198.38                    & 0.8862 \\
Retinopathy       & 22650.04                    & 0.8372 \\
\bottomrule
\end{tabularx}
\label{table:feature_map_comparison}
\end{table}

To quantify the impact of visual impairments on feature representations, feature maps were extracted from the final convolutional block (layer4) of each model. Table~\ref{table:feature_map_comparison} summarizes two complementary metrics relative to the normal model. \textbf{Disorder Activation Energy} measures the aggregate magnitude of activations, capturing the overall response strength of the network under degraded inputs. It is defined as the sum of absolute activations across spatial dimensions and channels:

\[
E = \sum_{i,j,k} \left| A_{i,j,k} \right|
\]

where \(A_{i,j,k}\) is the activation value at spatial location \((i, j)\) and channel \(k\). Deviations from the normal baseline energy (23807.8086) indicate changes in excitation levels caused by impaired inputs; higher values often reflect compensatory activation to recover salient structure, whereas lower values can imply attenuated feature responses. \textbf{Cosine Similarity} measures alignment between disorder-specific and normal feature maps by computing the angular similarity of their flattened representations. High similarity indicates preserved feature geometry, while low similarity indicates substantial representational drift. Together, these metrics characterize both magnitude and directional shifts in the learned feature space.

The analysis reveals that cataract and glaucoma have the most disruptive effects on feature representations, as evidenced by their low cosine similarity values (0.6350 and 0.4551). In contrast, refractive errors and retinopathy maintain higher similarity (0.8862 and 0.8372), indicating relatively preserved feature geometry. The heatmaps in Figure 3 corroborate these trends: cataract and glaucoma yield pronounced deviations in high-frequency regions, consistent with blurring or peripheral field loss.

The \textbf{cataract} model exhibits the highest disorder activation energy and a cosine similarity of 0.6350, indicating substantial deviations from normal feature maps. This shows how cataracts impair edge detection and contrast sensitivity in the primary visual cortex (V1), leading to degraded downstream processing in higher-order visual areas like the fusiform face area (FFA) \cite{b10}. The model’s difficulty in maintaining normal feature representations reflects the critical role of sharp visual input for effective recognition.

\textbf{Glaucoma} results in the lowest cosine similarity (0.4551), reflecting severe disruptions in feature alignment. This parallels the effect of glaucoma-induced peripheral vision loss on cortical representations, where the lack of peripheral input leads to incomplete spatial encoding in V1 and associated areas.\cite{b11}

The \textbf{AMD} model retains relatively high cosine similarity (0.9344), suggesting moderate disruption to feature maps. This corresponds to the loss of central vision in AMD, which predominantly affects regions of the visual cortex specialized for high-acuity tasks \cite{b12}. The preserved activation energy suggests that peripheral visual input compensates for central deficits, a phenomenon observed in the \textit{neural adaptation} mechanisms of AMD patients.

\textbf{Refractive errors} show minor disruptions, with a cosine similarity of 0.8862 and a modest increase in activation energy. This reflects how the brain can compensate for blurriness through top-down processing and contextual integration, allowing higher-order visual areas to maintain functionality despite input distortions \cite{b13}. The computational model aligns with this resilience, exhibiting minimal deviations in feature representations.

\textbf{Retinopathy} results in a cosine similarity of 0.8372 and a decrease in activation energy, indicating minimal but distinct disruptions. The scattered nature of retinal damage in retinopathy likely allows the brain to integrate information across unaffected regions \cite{b14}, maintaining relatively stable feature maps.

These findings demonstrate the model's sensitivity to input distortions and align with the known effects of these disorders on human visual processing.

\section{Conclusion}
\label{conclude}
This study presents a comprehensive analysis of how visual impairments affect feature representations in computational models. By leveraging the \textbf{BrokenEyes} framework to simulate these disorders, the results reveal significant disruptions in feature map activations and classification confidence, particularly for cataract and glaucoma, which align with the known effects of these disorders on human neural processing. These findings underscore the potential of computational models to mimic and study the effects of visual impairments, offering insights into the interplay between degraded inputs and feature representation. The findings of this paper can guide the development of assistive AI systems designed to adapt to impaired visual inputs, improving accessibility for individuals with vision disorders. Additionally, integrating neuroscientific validation through fMRI or eye-tracking studies can further bridge the gap between computational simulations and human visual processing.

\end{document}